# Equitable Survival Prediction: A Fairness-Aware Survival Modeling (FASM) Approach

Mingxuan Liu, MSc[1,2], Yilin Ning, PhD[1,2], Haoyuan Wang[3], Chuan Hong[3], Matthew Engelhard[3,4], Danielle S. Bitterman, MD[5], William G. La Cava, PhD[6], Nan Liu, PhD[1,2,7,8]

[1] Centre for Quantitative Medicine, Duke-NUS Medical School, Singapore

[2] Duke-NUS AI + Medical Sciences Initiative (DAISI), Duke-NUS Medical School, Singapore

[3] Department of Biostatistics and Bioinformatics, Duke University School of Medicine, Durham, NC, USA

[4] Duke AI Health, Durham, NC, USA

[5] Artificial Intelligence in Medicine Program, Mass General Brigham, Harvard Medical School, Boston, MA, USA

[6] Computational Health Informatics Program, Boston Children's Hospital; Department of Pediatrics, Harvard Medical School, Boston, MA, USA

[7] Programme in Health Services and Systems Research, Duke-NUS Medical School, Singapore

[8] NUS Artificial Intelligence Institute, National University of Singapore, Singapore

# Abstract

As machine learning models become increasingly integrated into healthcare, structural inequities and social biases embedded in clinical data can be perpetuated or even amplified by data-driven models. In survival analysis, censoring and time dynamics can further add complexity to fair model development. Additionally, algorithmic fairness approaches often overlook disparities in cross-group rankings, e.g., high-risk Black patients may be ranked below lower-risk White patients who do not experience the event of mortality. Such misranking can reinforce biological essentialism and undermine equitable care. We propose a Fairness-Aware Survival Modeling (FASM), designed to mitigate algorithmic bias regarding both intra-group and cross-group risk rankings over time. Using breast cancer prognosis as a representative case and applying FASM to SEER breast cancer data, we show that FASM substantially improves fairness while preserving discrimination performance comparable to fairness-unaware survival models. Time-stratified evaluations show that FASM maintains stable fairness over a 10-year horizon, with the greatest improvements observed during the mid-term of follow-up. Our approach enables the development of survival models that prioritize both accuracy and equity in clinical decision-making, advancing fairness as a core principle in clinical care.

**Keywords**

AI fairness, fairness in healthcare, interpretable machine learning, survival analysis, trustworthy machine learning

# Introduction

As machine learning (ML) gains prominence in high-stakes fields like healthcare, concerns about bias have intensified.[1-3] Models trained on real-world data may perpetuate existing health disparities and further introduce algorithmic bias, particularly when demographic groups differ in care access, treatment quality, or follow-up.[4,5] In oncology, breast cancer remains a salient example, where despite advances in screening, diagnosis, and treatment for breast cancer, substantial disparities persist in breast cancer outcomes across sociodemographic groups.[6,7] Black women, in particular, are less likely to receive timely screening and more likely to receive delayed or lower-quality treatment, which contributes to shorter survival times compared with White women.[8-10]

In survival modeling, which typically estimates patient-specific risks over time, the temporal nature of outcomes adds complexity to algorithmic bias. Due to the long follow-up period,[11] censoring (i.e., loss to follow-up) often occurs unevenly across populations.[9,10,12] Marginalized groups, such as Black women, may experience higher censoring rates due to systemic barriers to consistent care.[6] This results in incomplete or biased training data, leading to inaccurate risk estimates and distorted hazard functions. Moreover, the impact of bias may change over time, making it essential to account for the temporal evolution of fairness in survival analysis.[13]

Conventional fairness-aware approaches aim to equalize performance within demographic subgroups, for example, making accuracy comparable between Black and White patients.[14,15] However, these intra-group evaluations often miss disparities in cross-group rankings—how individuals from one subgroup are ordered relative to those from another. In survival analysis, where outcomes include both event occurrence and survival times, such cross-group ranking disparities can be particularly harmful. For example, a model may accurately predict risk rankings within Black and White subgroups separately, yet still rank high-risk Black patients below lower-risk White patients.[16] This systematic misordering, or directional ranking bias, is especially concerning in clinical settings where treatment decisions or resource allocation depend on relative risk rankings.[17,18] Therefore, addressing ranking fairness, especially in a time-dependent manner, is critical to ensuring that patients are equitably prioritized for care over time.

In response to these challenges, we present a fairness-aware survival modeling (FASM) approach that explicitly accounts for disparities in time-dependent cross-group rankings. FASM can also navigate the balance between model performance and fairness by constructing diverse, nearly-optimal models that share similar predictive performance but with varying fairness profiles. In this study, we demonstrate FASM using breast cancer as a representative case due to its well-documented health disparities.[6-10,19] Applied to the Surveillance, Epidemiology, and End Results (SEER) breast cancer dataset, FASM significantly mitigates algorithmic biases in both intra- and cross-group risk rankings over time. It matches the predictive performance of conventional survival models while significantly improving fairness, especially during the mid-term of the follow-up period. Our approach supports the development of clinical decision tools that promote not only predictive accuracy but also fairness, ensuring that risk estimates inform care in an equitable manner across patient populations.

## Method

### Study cohort

The SEER Program collects cancer incidence and survival data from population-based cancer registries covering around 45.9% of the U.S. population.[20] A case-listing session was performed to identify women diagnosed with a first primary in situ or invasive breast cancer, using the SEER 8 registries based on the November 2023 submission. The dataset spanned 1975 to 2021. The reporting of this study followed the guidelines of TRIPOD+AI (Transparent Reporting of a multivariable prediction model for Individual Prognosis Or Diagnosis).[21] Because this data set is in the public domain, it was exempt from institutional review board approval and the requirement for informed consent.

To construct a clinically homogeneous cohort for fairness-aware survival analysis, we applied the following exclusion criteria. Patients younger than 21 years at breast cancer diagnosis were excluded to remove rare early-onset cases with distinct etiologies. Individuals who were not biologically female were excluded due to differences in disease biology and low sample sizes among male patients. We limited the cohort to Black and White individuals to ensure sufficient representation for comparative analysis. Patients were excluded if breast cancer was not their first primary malignancy or if the disease was classified as in situ (i.e., stage 0) or of unknown stage, to ensure inclusion of invasive, clinically significant cases with reliable staging information. Cases with missing tumor grade or unavailable cancer subtype were excluded to avoid confounding due to incomplete clinical profiles. We further excluded patients without a diagnosis of invasive ductal carcinoma (IDC), the most common and prognostically significant subtype of breast cancer[22], and those who did not undergo surgery[23], as its omission often reflects advanced or palliative contexts.

**Study variables**

Demographical characteristics (age, race/ethnicity, marital status, and residence location) and clinical characteristics (grade, stage, cancer subtype, radiation, and chemotherapy) were selected for modeling[7,24]. In the SEER database, marital status was dichotomized as married and unmarried groups, where the unmarried groups included single, separated, divorced, widowed, and unmarried or domestic partnership.[7] Residence location was categorized as metropolitan and nonmetropolitan. Tumor grade was grouped into low/intermediate (Grade I–II) and high grade (Grade III–IV). Cancer stage was categorized as early-stage (Stage I-II) or late-stage (Stage III-IV) due to their distinct differences in prognosis, treatment intent and clinical outcomes.[7,25] Radiation therapy (RT) was defined as receipt of any of the following: beam radiation, radioactive implants, isotopes, or unspecified radiation, and both RT and chemotherapy were binarized as yes vs. no/unknown.

In this study, the event of interest was mortality due to breast cancer, and patients were censored at the date of last follow-up if the event had not occurred. The final dataset was randomly divided into non-overlapping training (70%), validation (10%), and testing (20%) sets, stratified by race and event status to preserve distributional balance across sets.

**Model development: The framework of FASM**

To address bias in survival prediction, we developed the framework of Fairness-Aware Survival Modeling (FASM), which identifies models that achieve high predictive performance while minimizing bias. FASM consists of two components: generating a set of near-optimal models and selecting among them using fairness-aware criteria.

*Generation of nearly-optimal survival models*

In ML-based risk prediction, multiple models may achieve comparable levels of predictive performance while differing in their reliance on specific variables or subpopulation behaviors.[26-28] This variability is captured in what we refer to as the Rashomon set—a group of near-optimal models that allow exploration of fairness-performance trade-offs.[29-31] We extended this concept to survival modeling and focused on Cox proportional hazards (CoxPH) models for their interpretability and clinical relevance. The optimal model is defined as the full CoxPH model that maximizes partial likelihood.

To define the near-optimality, we used an adapted pseudo-$R^2$ measure ($R_L^2$), which is tailored for right-censored data for model performance measurement.[32] Unlike concordance-based measures such as Harrell's concordance index ($C$-index)[33], $R_L^2$ offers variance decomposition and reduces to classical $R^2$ in uncensored settings, making it more statistically interpretable. Notably, $R_L^2$ does not necessarily require a correctly specified model[32], which makes it well-suited for evaluating the set of nearly-optimal models. Models with predictive performance within a pre-specified margin of the optimal model were included in the Rashomon set. Candidate models were generated via rejection sampling, which perturbs the coefficients of the case-specific optimal model to explore near-optimal solutions based on validation-set performance.[29-31] Full methodological details on nearly-optimal model generation are provided in Supplementary eMethods.

We considered Rashomon sets at two levels to capture different patterns of variable reliance under varying inclusion/exclusion of sensitive variables. The sensitive variable-specific Rashomon set comprised near-optimal models constructed under fixed variable collections (e.g., with race included or excluded). The integral Rashomon set was the union of all case-specific sets. To maintain overall near-optimality, the pre-defined margin for each case-specific set ($\epsilon_0$) is stricter than that for the integral set ($\epsilon$), i.e., $\epsilon_0 < \epsilon$. In our previous studies about generating nearly-optimal models using Rashomon sets, $\epsilon$ is often set at 5%.[29-31]

*Model selection with fairness*

To prioritize fairness among nearly-optimal models, we developed a Model Selection Index (MSI), which integrates multiple fairness metrics into a single composite score.[31] See more details about quantitative fairness in the next subsection, "Fairness evaluation". Inspired by the radar chart for multidimensional comparisons[31], MSI is a holistic ranking measure that accounts for not only individual bias dimensions ($m_1, m_2, ..., m_J$) but also their interdependencies, calculated as:

$$\text{MSI}(f_\beta) = \frac{1}{\sum_{j=1}^{J} m_j(f_\beta) m_{j+1}(f_\beta)},$$

where $m_{J+1} := m_1$ for simplicity. The model with the highest MSI score within the Rashomon set is chosen as the final FASM model (i.e., the fairness-aware model).

**Fairness evaluation**

From the set of nearly-optimal models, we identified a fairer one considering both intra- and cross-group ranking fairness (Table 1). The fairness metrics are derived from performance measures used to evaluate risk rankings. Discrepancies in ranking performance were interpreted as indicators of biases, i.e., a lack of fairness. Primary performance metrics employed in this study included:

$C$-index: Harrell's $C$-index[33] quantifies the probability that, for a randomly selected pair of comparable individuals, the model assigns a higher risk score to the individual who experiences the event earlier.

Integrated $AUC$ ($iAUC$): The integrated $AUC$ ($iAUC$) summarizes model discrimination over time by averaging the time-dependent $AUC$, denoted $AUC(t)$, across the follow-up time period. At each time point, $AUC(t)$ represents the probability that the model correctly ranks an individual who experiences the event before $t$ higher than one who does not.[17,34]

Cross concordance index ($xCI$): This metric assesses whether the model correctly ranks individuals from one subgroup relative to individuals from another subgroup based on their observed event times.[16] For any subgroup pairs, e.g., subgroups $a$ and $b$, the $xCI_{(a,b)}$ is calculated by finding all comparable pairs of individuals $i$ and $j$ from subgroups $a$ and $b$ respectively, where (1) individual $i$ belongs to subgroup $a$ had an observed event at some time $t_i$; and (2) individual $j$ belonging to subgroup $b$ had an event later or was censored at $t_j > t_i$.

cross $AUC$ ($xAUC$): As a time-specific extension of $xCI$, $xAUC_{(a,b)}(t)$ estimates the probability that an individual from the subgroup $a$ who experiences the event before time $t$ is ranked higher than an individual from the subgroup $b$ who experiences the event after time $t$ or not at all, using ROC-based discrimination.[34]

To address censoring, we incorporated inverse probability of censoring weights (IPCW)[35] to estimate the metrics. IPCW corrects for informative censoring by reweighting observed events according to the probability of remaining uncensored,[35] providing a valid basis for both intra-group and cross-group fairness measures.

*Intra-group ranking bias*

The FASM framework first emphasizes the equality of model performance among subgroups.[31] Analogous to equal opportunity in binary classification, which emphasizes the equality of true positive rates among subgroups, we assess fairness in survival models using $C$-index and $iAUC$. Disparities in these metrics across subgroups indicate bias, referring to intra-group ranking bias. These are captured by:

Disparity in $C$-index ($\Delta CI$): Maximum absolute difference in $C$-index among all subgroups.

Disparity in $iAUC$ ($\Delta iAUC$): Maximum absolute difference in $iAUC$ among all subgroups.

Lower values of $\Delta CI$ and $\Delta iAUC$ reflect a lower bias and better fairness. $\Delta CI$ and $\Delta iAUC$ can be highly correlated if the risk rankings remain stable over time. When rankings vary across time, such as in the case of breast cancer[36], $\Delta iAUC$ can diverge from $\Delta CI$ and add additional information about fairness over time.

*Cross-group ranking bias*

Intra-group ranking bias assesses how individuals are ranked within their own subgroup but can not capture disparities in how individuals are ranked relative to those in other subgroups. As a result, a model may correctly assign higher risk scores to event cases within a disadvantaged group, yet still assign them lower absolute risk scores than non-events in a more privileged group.[17] This cross-group misranking introduces systematic bias that intra-group metrics alone fail to detect. To capture this, we considered the cross-group ranking bias, measured based on $xCI$ and $xAUC(t)$, with:

Disparity in $xCI$ ($\Delta xCI$): A fair model is expected to yield symmetric rankings across subgroups, i.e., $xCI_{(a,b)} = xCI_{(b,a)} \ \forall a \neq b$. We define the disparity regarding $xCI$ with the maximum absolute difference between all reciprocal $xCIs$, noted as $\Delta xCI$. A smaller $\Delta xCI$ indicates more consistent cross-rankings and thus fairer model behavior.

Integral disparity in $xAUC$ ($i\Delta xAUC$): As a time-specific extension of $\Delta xCI$, we define the disparity regarding $xAUC(t)$ with the maximum absolute difference between all reciprocal $xAUC(t)$ values, noted as $\Delta xAUC(t)$. To aggregate the disparities measured by $\Delta xAUC(t)$ over time, we introduced $i\Delta xAUC$ to summarize the maximum cross-group disparity in $xAUC$ values over a specified time interval. A smaller $ixAUC$ value indicates a lower cross-group ranking bias and greater fairness over time. Similarly, $\Delta xCI$ and $i\Delta xAUC$ can be correlated if the cross-group risk rankings remain stable over time; in the presence of time-varying cross-group ranking, $i\Delta xAUC$ can diverge from $\Delta xCI$, capturing additional temporal cross-group bias dynamics.

For clarity, we use the prefix $x$ to denote cross-group metrics, $i$ to indicate time-integrated metrics, and $\Delta$ to represent subgroup disparities in the corresponding performance metric. Both intra-group ($\Delta CI$, $\Delta iAUC$) and cross-group ($\Delta xCI$, $i\Delta xAUC$) ranking bias metrics were incorporated into the MSI calculation for fairness-aware model selection.

### Statistical analysis

In the descriptive analysis of variables of interest, continuous variables were summarized by mean and standard deviation as well as median and interquartile range, while categorical variables were summarized by frequency and percentage. Comparisons between groups (White vs Black) were performed using the Chi-square test for categorical variables and the Mann-Whitney U test for continuous variables after the Kolmogorov-Smirnov test verified non-normality.

We compared the proposed fairness-aware survival model ("FASM") with two CoxPH baselines: a full model including race ("CoxPH") and a model excluding race ("Under-blindness"). We assessed model performance using standardized metrics of survival modeling $C$-index and $iAUC$, with 95% confidence intervals (CI) estimated via bootstrapping. We evaluated overall model fairness using both intra-group ranking measures ($\Delta CI$ and $\Delta iAUC$) and cross-group ranking measures ($\Delta xCI$ and $i\Delta xAUC$). In addition, we assessed the time dynamics of model fairness using $\Delta xAUC(t)$ between White and Black subgroups. The data analysis and model building were performed using R version 4.0.2 (The R Foundation for Statistical Computing) and Python version 3.9.7.

## Results

### Patient demographic and clinical characteristics

Among 463,938 breast cancer patients identified in the SEER database, 47,618 patients (10.3%) met the eligibility criteria for this study, of whom 4,747 (10.0%) were Black and 42,871 (90.0%) were White (Table 2). As shown in eTable 1, exclusions were primarily due to missing or ineligible clinical data, including unknown cancer stage (35.2%), unavailable subtype information (31.7%), and non-Black or non-White race (11.5%). Additional exclusions included patients under 21 years of age, non-biological females, non-invasive histologies, and those who did not undergo surgery. As shown in eFigure 1, long-term survival probabilities were consistently lower for Black patients compared to White patients throughout the 10-year follow-up period.

Table 2 summarizes key demographic and clinical features stratified by race. The overall mean (SD) age at diagnosis was 59.5 (13.0) years, with Black patients being younger than White patients (56.0 vs. 59.9 years). Compared with White patients, Black patients were less likely to be married (36.0% vs. 60.2%), more likely to reside in metropolitan areas (99.2% vs. 85.8%), and more likely to present with higher-grade (52.7% vs. 33.6%) and late-stage disease (16.8% vs. 10.7%). Subtype distributions also varied significantly by race; for instance, triple-negative tumors were more common in Black patients (21.6% vs. 11.2%). Black patients were more likely to receive chemotherapy (62.0% vs. 43.1%), and had shorter average survival times (89.0 vs. 93.3 months) and higher event rates (14.4% vs. 8.1%).

**Nearly-optimal models and the FASM model**

Figure 1a visualizes the distribution of model coefficients across the Rashomon set, which was defined as the collection of models achieving within 1.05 times of $R_L^2$ of the optimal survival model (i.e., CoxPH). While all models in this set exhibit near-optimal performance, they vary in their reliance on specific covariates. Notably, coefficients for bias-related variables such as race, marital status, and metropolitan residence show substantial variability. This indicates that different models can encode differing levels of dependence on these attributes, despite similar overall predictive performance.

The FASM model selected from the Rashomon set was the one with the highest *MSI*. This model was race-free, derived from the case-specific Rashomon set that excluded race. As shown in Figure 1b, compared with the CoxPH model that was fairness-unaware, FASM downplayed social determinants such as marital status and metropolitan residence. For clinical variables, FASM displayed different coefficient patterns from the CoxPH model. FASM gave greater relative weight to HR+HER2+ compared with other subtypes and emphasized cancer stage while placing less weight on tumor grade. Unlike the CoxPH model, which suggested chemotherapy was associated with poorer survival, FASM treated chemotherapy as a protective factor linked to improved survival, though this association did not reach statistical significance.

**Model fairness and performance**

Table 3 compares fairness and performance across three survival modeling approaches. The FASM model demonstrated the best overall fairness, with the minimal intra-group ranking bias ($\Delta iAUC = 0.003$; $\Delta CI = 0.013$) and the smallest cross-group bias ($\Delta xCI = 0.132$; $i\Delta xAUC = 0.006$). In contrast, the CoxPH model exhibited higher bias, particularly regarding $\Delta xCI$ (0.261), indicating more pronounced cross-group ranking bias. The Under-blindness model reduced cross-group ranking bias slightly ($\Delta xCI = 0.163$) compared to CoxPH but introduced larger $i\Delta xAUC$ (0.043), suggesting temporal fluctuation in ranking fairness.

In terms of predictive performance, all models achieved comparable discrimination, with $iAUC$s ranging from 0.827 to 0.833 and $C$-indices from 0.758 to 0.766. While the FASM model had a marginally lower $iAUC$ (0.827; 95% CI: 0.815-0.839) than the CoxPH (0.833; 95% CI: 0.821-0.843), it achieved the most balanced trade-off between fairness and performance.

As a complement to $i\Delta xAUC$, which summarizes the disparity in $xAUC$ over time, Figure 2 shows the year-by-year disparity in time-dependent $xAUC$s, i.e., $\Delta xAUC(t)$, across a 10-year follow-up period. The FASM model consistently maintains low and stable cross-group ranking disparities over time, with $\Delta xAUC(t)$ remaining below 0.02 across all time points. In contrast, the CoxPH model exhibits a sharp disparity in the first year. The Under-blindness model exhibits worsening disparities over time, with $\Delta xAUC(t)$ of 0.068 at the end.

Figure 3 further illustrates fairness by comparing intra-group and cross-group $C$-indices. All three models showed comparable performance for both White and Black subgroups. However, substantial disparities emerged in cross-group settings. In particular, the CoxPH model demonstrated marked directional bias, with much lower $C$-index when comparing White cases to Black controls and disproportionately higher $C$-index for Black cases versus White controls. This directional bias indicates a systematic misranking of Black patients relative to White patients. In contrast, FASM notably reduced this cross-group ranking disparity, improving the fairness of risk rankings across racial subgroups without compromising intra-group predictive performance. The Under-blindness model also narrowed the disparity, though less effectively than FASM.

Figure 4 shows the distribution of predicted risks stratified by race over the follow-up years. Compared with the CoxPH and Under-blindness models, the FASM model yielded more balanced predictions between Black and White patients. The CoxPH model systematically assigned higher risk scores to Black patients throughout follow-up. FASM successfully reduced this discrepancy, yielding similar risk predictions across racial groups while maintaining the distinction between censored and event cases. The discrepancy was smallest between years 2-5, though risk differences between Black and White patients gradually re-emerged later in follow-up; nevertheless, these disparities remained smaller than those observed with CoxPH or Under-blindness models.

## Discussion

Disparities in breast cancer have been evident in previous literature; the development of a prediction model should not further exacerbate the disparities. In this work, we proposed a fairness-aware survival modeling approach for breast cancer survival prediction that accounts for racial disparities in time-dependent cross-group rankings. Our method mitigates disparities regarding both intra- and cross-group risk rankings over time, especially during the mid-term stages of follow-up. Our approach supports the development of clinical decision tools that promote equitable access to timely, life-saving care.

A key contribution of FASM lies in its ability to address disparities in not only intra- but also cross-group risk rankings, which traditional models often overlook[16,37] (Figure 2-3). To quantify and mitigate this, we employed metrics $\Delta xCI$ and $i\Delta xAUC$ to evaluate cross-group ranking bias, extending the $xAUC$ metric in the binary classification[34] to survival models. These metrics estimate the probability that an individual from one group who experienced an event is correctly ranked above an individual from another group who experienced the event later or not at all. In a fair model, this probability should be consistent across subgroup pairs—that is, $xCI$s should be approximately equal for all group combinations. Our findings show that conventional survival models, which do not explicitly account for potential bias, often violate this criterion, but FASM substantially reduces these disparities (Table 2 and Figure 2-3), enhancing fairness in risk-based clinical prioritization.

Because individual fairness metrics capture different aspects of survival modeling[38,39] (Table 1), consistent gains across multiple measures are necessary to reliably assess a model's fairness. For example, compared with the CoxPH model, the Under-blindness model showed lower $\Delta xCI$ (Table 3), indicating the reduced overall disparity across the entire follow-up. However, it exhibited higher $\Delta xAUC$ at year 10 (Figure 3), reflecting larger disparities at the end horizon, as well as higher $i\Delta xAUC$ (Table 3), reflecting the greater average time-specific disparity over time. In contrast, FASM achieved lower disparities across these and other measures, demonstrating more robust fairness throughout the follow-up period.

Our results reveal that racial disparities in predicted breast cancer survival risks are dynamic rather than static, with similar temporal patterns across models. For FASM, risk predictions for Black and White patients were comparable in the middle-term (years 2-5) but diverged in the early (0-2 years) and late (up to 10 years) follow-up periods (Figure 4). The early disparity likely reflects the later-stage diagnosis among Black women, driven by unequal access to screening and diagnostic delays.[8] Because the late stage strongly predicts early mortality, survival gaps are pronounced in the initial years. Over longer follow-up, structural inequities—including interruptions in care, lower treatment adherence, limited access to new therapies, and broader socioeconomic challenges—compound survival gaps over time.[6,40] While fairness-aware models like FASM can help correct for algorithmic bias and promote more balanced risk predictions in the mid-term, they cannot fully counteract inequities rooted in healthcare delivery.[41] Our findings underscore the need to integrate algorithmic solutions with systemic interventions to address the underlying drivers of disparity.

Rather than directly removing race from the model, which can obscure structural inequities[42], we adopt a data-driven approach that evaluates whether sensitive variables should be included or excluded based on the fairness of the resulting models. Using Rashomon-set analysis, we explore a spectrum of nearly-optimal models and identify those that reduce disparities while preserving predictive performance. This strategy allows us to develop models that are not only fair but also suitable for equitable deployment in real-world clinical settings.[4,31]

FASM offers adaptability to other machine learning models (e.g., neural networks) in the future. For these models, post-hoc explanation methods could be applied to enhance their interpretability. Notably, the fine-grained temporal resolution (e.g., monthly intervals over 10 years) increases computational demands, especially for models with large parameter spaces.[13] Additionally, extending FASM to different architectures requires additional methodological adjustments, as coefficient perturbation may not be directly applicable.[27] In such cases, empirical strategies, such as applying random masks to neural network weight matrices, can be used to explore fairness-accuracy trade-offs.[43]

While FASM was demonstrated using breast cancer as a case study, its underlying framework is broadly applicable to other diseases characterized by survival outcomes and demographic disparities. The methodology is agnostic to disease type and can be adapted to any clinical context where survival modeling is relevant and fairness across subgroups is a concern. By accounting for both intra- and cross-group ranking disparities over time, FASM provides a generalizable approach for developing equitable survival models across diverse patient populations.

This work emphasizes ranking-based discrimination metrics, both intra- and cross-group, as the primary lens for fairness. While this focus supports equitable prioritization, we acknowledge that calibration—the agreement between predicted risks and observed outcomes—remains an important and complementary fairness consideration.[44] Despite the known inherent trade-off between calibration and discrimination-based fairness,[45] future work could integrate calibration-based fairness metrics to enable a more comprehensive assessment of model fairness.

This study has several limitations. First, our analysis focused on Black and White patients to enable a clearer assessment of disparities between the two largest subgroups, which may narrow the scope. Second, we limited the cohort to patients with IDC who underwent surgery, in order to ensure clinical homogeneity and reduce treatment-related confounding. This choice may exclude patients with other breast cancer subtypes or those receiving non-surgical management, limiting the applicability of our findings. Third, the lack of granular treatment information—particularly for therapies received after initial treatment—restricts our ability to adjust for differences in care quality. Finally, although the dataset is large, it spans multiple decades, during which breast cancer treatment has evolved considerably, introducing potential temporal heterogeneity.

# Conclusion

In this study, we introduced a fairness-aware survival modeling approach that accounts for disparities in time-dependent, cross-group risk rankings. FASM improves both intra-group and cross-group ranking fairness while maintaining strong predictive performance. Applied to SEER breast cancer data, FASM notably reduced racial disparities in risk predictions, particularly in mid-term follow-up periods. These findings underscore the importance of fairness in clinical risk modeling and offer a practical pathway toward more equitable AI deployment in clinical settings.

**Ethics declarations**

Ethics approval and consent to participate: Not applicable.

## Author Contributions

Conceptualization: M.L., N.L. Method design, experiment, data analysis, drafting of the manuscript: M.L. Critical revision of the manuscript: M.L., Y.N., H.W., C.H., D.S.B., W.G.L., N.L. Interpretation of the content: M.L., Y.N., H.W., C.H., M.E., D.S.B., W.G.L., N.L. Revisions of the manuscript: M.L., Y.N., H.W., C.H., M.E., D.S.B., W.G.L., N.L. Final read and approval of the completed version: all authors. Overseeing the project: N.L.

## Acknowledgement

This work was supported by the Duke-NUS Signature Research Programme funded by the Ministry of Health, Singapore. Any opinions, findings and conclusions or recommendations expressed in this material are those of the author(s) and do not reflect the views of the Ministry of Health.

## Competing Interests

The authors declare that there are no competing interests.

**Table 1.** Summary of performance and fairness metrics

| Performance metric[1,2,3] | Description | Fairness metric[1,2,3] | Description |
|---|---|---|---|
| $C$-index | Overall ranking ability: measures whether an individual with an earlier event is assigned a higher risk score than one with a later event or not at all, over the entire follow-up period. | $\Delta CI$ | Maximum discrepancy in $C$-index across subgroups. |
| $iAUC$ | Time-integrated discrimination: averaging time-specific discrimination $AUC(t)$ across the follow-up period. | $\Delta iAUC$ | Maximum discrepancy in $iAUC$ across subgroups. |
| $xCI_{(a,b)}$ | Overall cross-group ranking: measures whether an individual $i$ in subgroup $a$ with an earlier event is ranked above individual $j$ in subgroup $b$ with a later event or not at all, over the entire follow-up time period. | $\Delta xCI$ | Maximum absolute difference in reciprocal $xCI$ values (e.g., $xCI_{(a,b)}$ and $xCI_{(b,a)}$) across all subgroup pairs. |
| $xAUC_{(a,b)}(t)$ | Time-specific cross-group discrimination: measures whether an individual $i$ in subgroup $a$ with an event before time $t$ is ranked above individual $j$ in subgroup $b$ with an event after time $t$ or not at all. | $\Delta xAUC(t)$ | Maximum absolute difference in reciprocal $xAUC$ values at time $t$ across all subgroup pairs. |
| | | $i\Delta xAUC$ | Time-integrated disparity, averaging disparities measured by $\Delta xAUC(t)$ across the follow-up period. |

[1] The prefix "$x$" denotes cross-group metrics; "$i$" indicates time-integrated metrics; "$\Delta$" represents disparities regarding the corresponding performance metric. Lower Δ values indicate smaller disparities and greater fairness.
[2] All estimates were adjusted for censoring using inverse probability of censoring weights (IPCW).
[3] More mathematical details are provided in Supplementary eTable 2.

**Table 2.** Patient demographic and clinical characteristics by race

|  | Overall (n=47,618) | White (n=42,871) | Black (n=4,747) |
|---|---|---|---|
| Age, mean (SD), y | 59.5 (13.0) | 59.9 (13.0) | 56.0 (12.7) |
| **Marital Status, n (%)** | | | |
| No/Unknown | 20115 (42.2) | 17075 (39.8) | 3040 (64.0) |
| Yes | 27503 (57.8) | 25796 (60.2) | 1707 (36.0) |
| **Metropolitan, n (%)** | | | |
| No/Unknown | 6109 (12.8) | 6069 (14.2) | 40 (0.8) |
| Yes | 41509 (87.2) | 36802 (85.8) | 4707 (99.2) |
| **Grade, n (%)** | | | |
| IorII | 30713 (64.5) | 28468 (66.4) | 2245 (47.3) |
| IIIorIV | 16905 (35.5) | 14403 (33.6) | 2502 (52.7) |
| **Stage, n (%)** | | | |
| Early | 42213 (88.6) | 38264 (89.3) | 3949 (83.2) |
| Late | 5405 (11.4) | 4607 (10.7) | 798 (16.8) |
| **Subtype, n (%)** | | | |
| HR-HER2- | 5811 (12.2) | 4787 (11.2) | 1024 (21.6) |
| HR-HER2+ | 2307 (4.8) | 2004 (4.7) | 303 (6.4) |
| HR+HER2- | 34110 (71.6) | 31324 (73.1) | 2786 (58.7) |
| HR+HER2+ | 5390 (11.3) | 4756 (11.1) | 634 (13.4) |
| **Radiation, n (%)** | | | |
| No/Unknown | 17215 (36.2) | 15499 (36.2) | 1716 (36.1) |
| Yes | 30403 (63.8) | 27372 (63.8) | 3031 (63.9) |
| **Chemotherapy, n (%)** | | | |
| No/Unknown | 26184 (55.0) | 24382 (56.9) | 1802 (38.0) |
| Yes | 21434 (45.0) | 18489 (43.1) | 2945 (62.0) |
| Survival time, mean (SD) | 92.9 (28.4) | 93.3 (28.1) | 89.0 (30.9) |
| Event rate, n (%) | 4140 (8.7) | 3458 (8.1) | 682 (14.4) |

**Table 3.** Two-dimensional evaluation of model fairness and performance for the fairness-unaware CoxPH model, the Under-blindness model, and the FASM model

|  | **Fairness metrics ↓** | | | | **Performance metrics ↑** | |
|---|---|---|---|---|---|---|
|  | $\Delta iAUC$ | $\Delta CI$ | $\Delta xCI$ | $i\Delta xauc$ | $iAUC$ | $CI$ |
| CoxPH | 0.006 | 0.016 | 0.261 | 0.016 | 0.833 [0.821, 0.843] | 0.766 [0.753, 0.778] |
| Under-blindness | 0.007 | 0.016 | 0.163 | 0.043 | 0.833 [0.818, 0.845] | 0.765 [0.750, 0.779] |
| FASM | 0.003 | 0.013 | 0.132 | 0.006 | 0.827 [0.815, 0.839] | 0.758 [0.745, 0.771] |

**Figure 1.** Coefficients of nearly-optimal models within the Rashomon set (a) and coefficients with 95% confidence intervals for fairness-unaware (i.e., CoxPH) and fairness-aware (i.e., FASM) models (b)

**(a)**

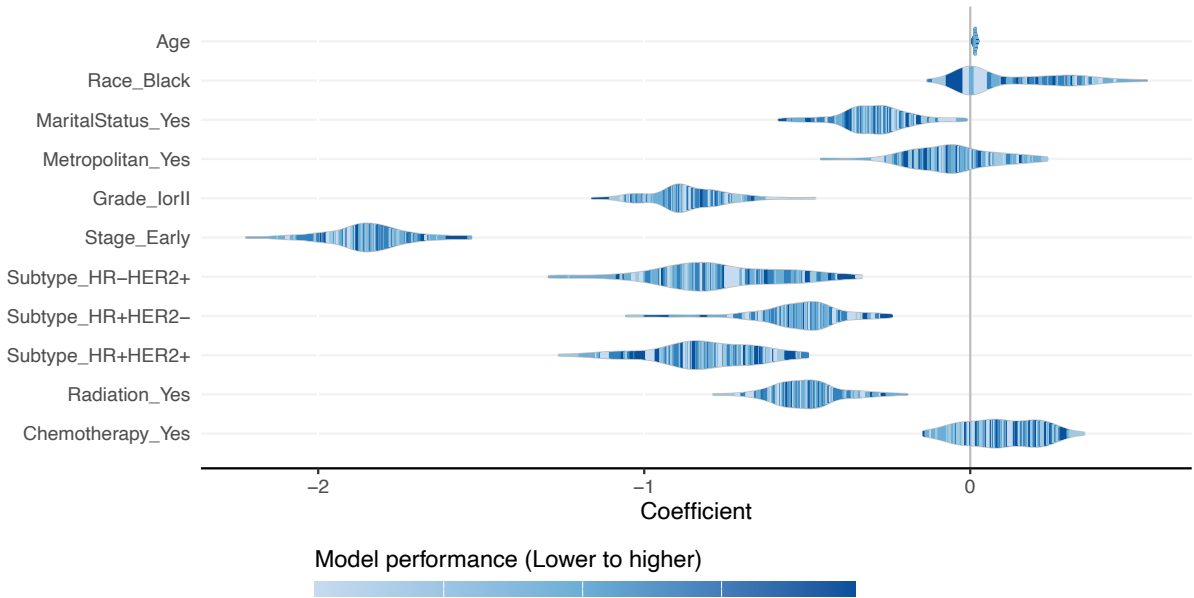

**(b)**

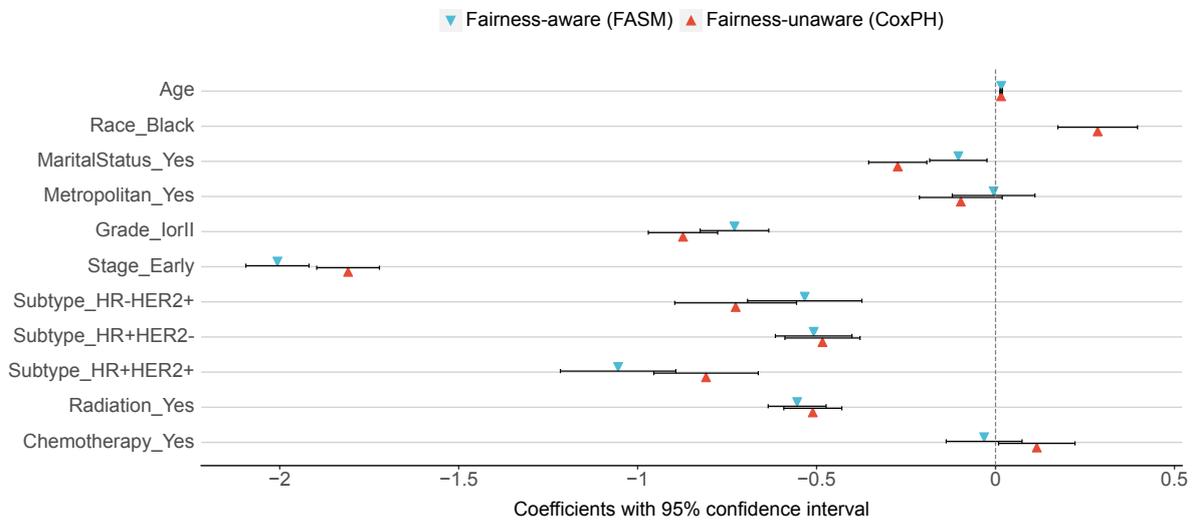

**Figure 2.** Comparison of time-dependent cross-group fairness disparities over a 10-year follow-up period across three models.

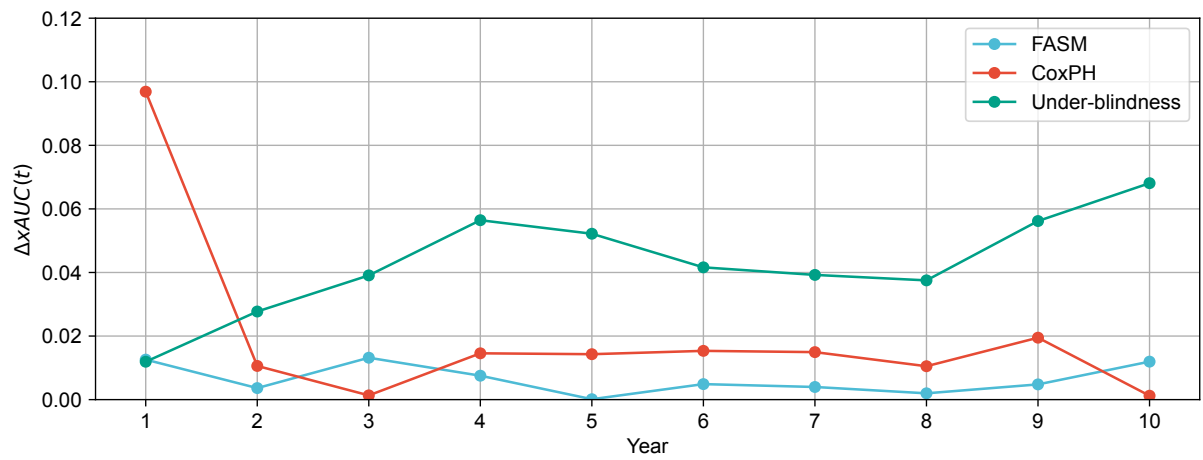

**Figure 3.** Comparison of intra-group and cross-group $C$-index across models.

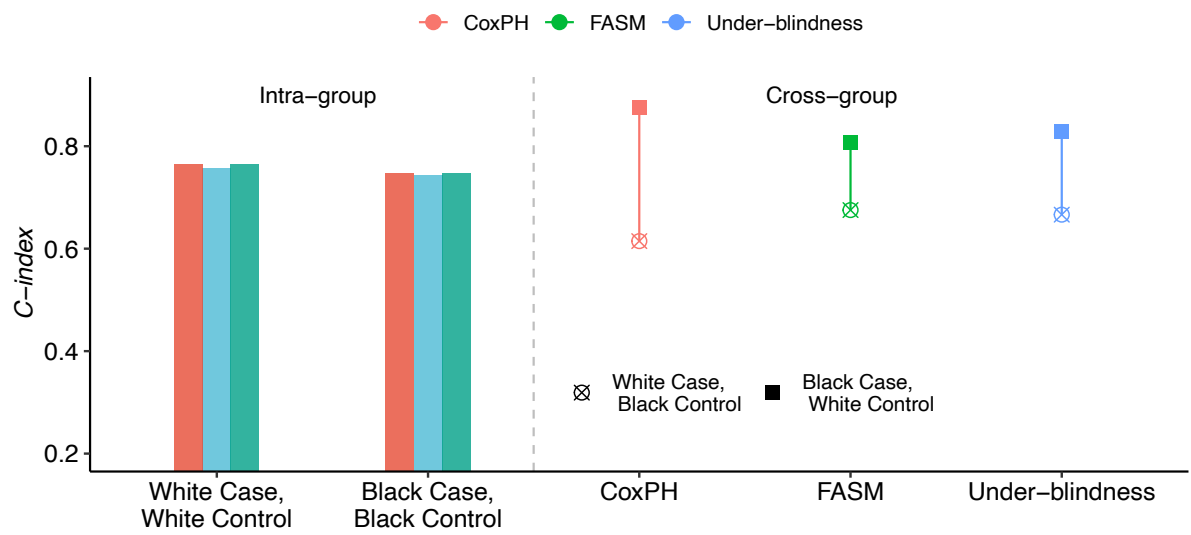

**Figure 4.** Comparison of predicted risk between models over 10 years

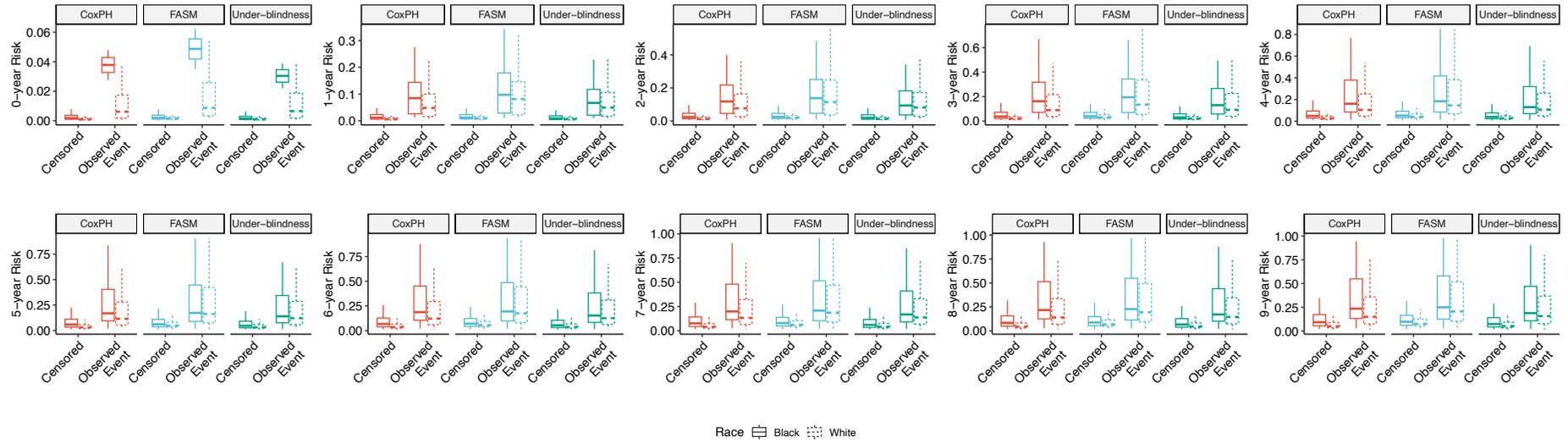

# Supplementary

**eTable 1.** General Exclusions

| Exclusion | N | % |
|---|---:|---:|
| Age <21 at breast cancer diagnosis | 54 | 0.01 |
| Not biological female | 3,187 | 0.69 |
| Not Black or White | 53,530 | 11.54 |
| Stage 0 or unknown stage cancer | 163,107 | 35.16 |
| Grade is unknown | 29,481 | 6.35 |
| Cancer subtype is not available | 147,026 | 31.69 |
| Not invasive ductal carcinoma | 17,072 | 3.68 |
| No surgery performed | 2,863 | 0.62 |
| *Eligible for study* | 47,618 | 10.26 |
| **Total** | 463,938 | 100 |

**eTable 2** Performanc and fairness metrics

| Metric | Type | Description | Equation |
|---|---|---|---|
| Concordance index ($C$-index)[1] | Performance metric | Probability that an individual with an earlier event is assigned a higher risk score than one with a later event or not at all, among comparable individual pairs, within the population. | $C\text{-index} = P\big(R(X_i) > R(X_j) \mid T_i < T_j\big)$ |
| $C\text{-index}_a$ | Performance metric | $C$-index in subgroup $a$. | $C\text{-index}_a = P\big(R(X_i) > R(X_j) \mid T_i < T_j, A_i = A_j = a\big)$ |
| $CI$ disparity ($\Delta CI$) | Fairness metric | Maximum absolute difference in $C$-index across all subgroup pairs. | $\Delta CI = \max\limits_{\substack{a,b \in G \\ a \neq b}} \lvert C\text{-index}_a - C\text{-index}_b \rvert$ |
| $xCI_{(a,b)}$[1] | Performance metric | Probability that an individual $i$ in subgroup $a$ with an earlier event is ranked above individual $j$ in subgroup $b$ with a later event or not at all, among all comparable pairs in subgroups $a$ and $b$. | $xCI_{(a,b)} = P\big(R(X_i) > R(X_j) \mid T_i < T_j, A_i = a, A_j = b\big)$ |
| $xCI$ disparity ($\Delta xCI$) | Fairness metric | Maximum absolute difference in $xCI$ across all subgroup pairs. | $\Delta xCI = \max\limits_{\substack{a,b \in G \\ a \neq b}} \lvert xCI_{(a,b)} - xCI_{(b,a)} \rvert$ |
| $iAUC(t_1, t_2)$ | Performance metric | Time-integrated AUC over interval $(t_1, t_2)$ within the population population, over time-period $(t_1, t_2)$. | $AUC(t) = E_{X_i\mid Y_i(t)=1} E_{X_j\mid Y_j(t)=0}\big[1(R(X_i) > R(X_j))\big]$ $iAUC(t_1, t_2) = \dfrac{1}{\hat{S}(t_1) - \hat{S}(t_2)} \int_{t_1}^{t_2} AUC(t)\, d\hat{S}(t)$ |
| $iAUC_a(t_1, t_2)$[2] | Performance metric | $iAUC(t_1, t_2)$ in subgroup $a$. | $AUC_a(t) = E_{X_i\mid Y_i(t)=1, A_i=a} E_{X_j\mid Y_j(t)=0, A_j=a}\big[1(R(X_i) > R(X_j))\big]$ |
| | | | $iAUC_a(t_1, t_2) = \dfrac{1}{\hat{S}(t_1) - \hat{S}(t_2)} \int_{t_1}^{t_2} AUC_a(t) d\hat{S}(t)$ |
| $iAUC$ disparity ($\Delta iAUC$) | Fairness metric | Maximum absolute difference in $iAUC$ across subgroup pairs. | $\Delta iAUC = \max\limits_{\substack{a,b \in G \\ a \neq b}} \|iAUC_a - iAUC_b\|$ |
| $xAUC_{(a,b)}(t)$ | Performance metric | Time-specific probability that that an individual $i$ in subgroup $a$ with an event before time $t$ is ranked above individual $j$ in subgroup $b$ with an event after time $t$ or not at all, among all comparable pairs at time $t$ in subgroups $a$ and $b$. | $xAUC_{(a,b)}(t) = E_{X_i\|Y_i(t)=1, A_i=a} E_{X_j\|Y_j(t)=0, A_j=b} [1(R(X_i) > R(X_j))]$ |
| $i\Delta xAUC(t_1, t_2)^2$ | Fairness metric | Maximum cross-group disparity in $xAUC$ over time-period $(t_1, t_2)$. | $ixAUC(t_1, t_2) = \dfrac{1}{\hat{S}(t_1) - \hat{S}(t_2)} \int_{t_1}^{t_2} \max\limits_{\substack{a,b \in G \\ a \neq b}} \|xAUC_{(a,b)}(t) - xAUC_{(b,a)}(t)\| \, d\hat{S}(t)$ |

[1] We used the Inverse Probability of Censoring Weighting (IPCW) to make the estimate more unbiased to censoring.

[2] $\hat{S}(t)$ is the estimated survival function for the overall population.

**eFigure 1.** Survival curves stratified by race

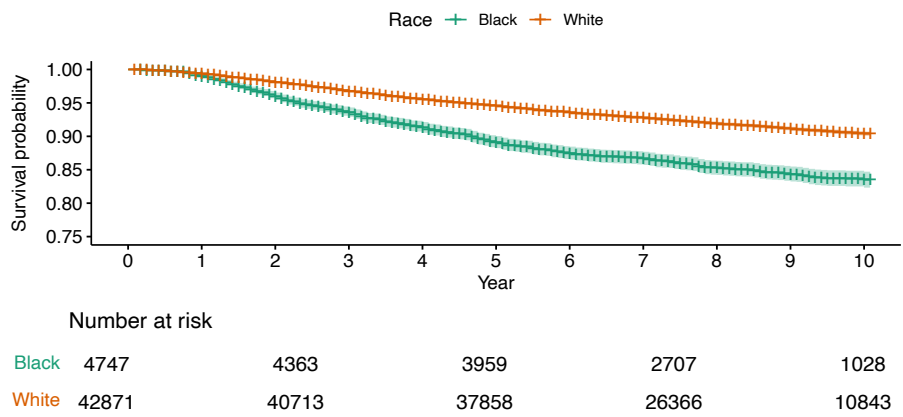

**eMethods.** Rashomon set, $R_L^2$ measure, Rejection sampling

- Rashomon set

Let $f^*_{(U,S)}$ denote the optimal model that maximizes the partial likelihood in the model family $F_{(U,S)}$, constructed using less-sensitive variables $X_U$ and sensitive variable $X_S$. $X_U$ typically include clinical variables such as cancer stage and grade. When building a model that excludes sensitive variables, such as a race-free model, $S$ is an empty set.

The $S$-specific Rashomon set is defined as:

$$R_{(U,S)}(\epsilon_0, \beta^*_{(U,S)}, B_{(U,S)}) = \{\beta \in B_{(U,S)} \mid R_L^2(f_\beta, Y) \geq (1-\epsilon_0) R_L^2(f^*_{(U,S)}, Y)\},$$

where "near-optimality" is controlled by the small factor $\epsilon_0 > 0$, $B_{(U,S)}$ is the coefficient space of models in $F_{(U,S)}$, and $\beta^*_{(U,S)}$ is the coefficient of the optimal model $f^*_{(U,S)}$.

- $R_L^2$ measure: combination of $R^2$ and $L^2$:

The measure $R^2$ proposed by Guo et al.[32] is inspired by the classical $R^2$ in linear regression but tailored to accommodate right-censored data without requiring a correctly specified model. $R^2$ is defined as the proportion of explained variance by a linearly corrected prediction model, quantifies the potential predictive power of the nonlinear prediction model.[32] The second measure $L^2$, defined as the proportion of explained prediction error by its corrected prediction function, gauges the closeness of the prediction function to its corrected version and serves as a supplementary measure to indicate (by a value less than 1) whether the correction is needed to fulfill its potential predictive power and quantify how much prediction error reduction can be realized with the correction. The two measures together provide a complete summary of the predictive accuracy of the nonlinear prediction function. We constructed $R_L^2 = wR^2 + (1-w)L^2$ to combine the effects of $R^2$ and $L^2$, where $w$ is the weightage for $R^2$ with a default value of 0.5.

- Experimental procedures of rejection sampling

To objectively evaluate the impact of sensitive variables on model performance and fairness, we included different cases of variable selection for $X_{S'}$. These include full inclusion (baseline: $X_{(U,S)}$), complete exclusion ("Under blindness": $X_U$) and all possible partial exclusions. We defined the integral Rashomon set as the union of case-specific near-optimal model sets, ensuring that all combinations of sensitive variable inclusion are represented:

$$R\left(\epsilon, \beta^*_{(U,\cdot)}, B_{(U,\cdot)}\right) = \bigcup_{\substack{S' \subseteq S \\ \beta^*_{(U,S')} \in R_{(U,S)}\left(\epsilon_0, \beta^*_{(U,S)}, B_{(U,S)}\right)}} R_{(U,S')}\left(\epsilon_0, \beta^*_{(U,S')}, B_{(U,S')}\right).$$

To maintain overall near-optimality determined by $\epsilon$, the case-specific near-optimality determined by $\epsilon_0$ should be more stringent, i.e., $\epsilon > \epsilon_0 > 0$. In previous studies, $\epsilon$ is typically set at 5%.[29,30]

We used rejection sampling[30,46] to generate the nearly-optimal models within each case-specific Rashomon set. Specifically, the $i$-th coefficient vector is generated from a multivariable normal distribution $N\left(\beta^*_{(U,S')}, k_i \Sigma^*_{(U,S')}\right)$ centered at the optimal coefficient $\beta^*_{(U,S')}$ with variance-corvariance matrix $k_i \Sigma^*_{U,S'}$. Here, $k_i$ is randomly drawn scaling factor from a uniform distribution $U(u_1, u_2)$ with tunable parameters $u_1$ and $u_2$ to adjust the scope of sampling. The baseline function is inherited from $f^*_{(U,S)}$. Sampling guided by the characteristics of the optimal model enables the efficient generation of nearly-optimal models and addresses model diversity within near-optimality of performance. This targeted sampling approach efficiently explores the space of nearly-optimal models, capturing variations in coefficients and fairness profiles while preserving performance, as illustrated in Figure 1.